\documentclass[acmsmall]{acmart}

\usepackage{amsmath}
\usepackage{lineno,hyperref}
\usepackage{algorithm}
\usepackage{algpseudocode}
\usepackage{bbding}
\usepackage{pifont}
\usepackage{wasysym}
\usepackage{amssymb}

\usepackage{amssymb}
\usepackage{booktabs}
\usepackage{multicol}
\usepackage{multirow}
\usepackage{diagbox}

\usepackage{longtable} 
\usepackage{booktabs}
\usepackage{graphicx}
\usepackage{caption}
\usepackage{subfigure}
\usepackage{float}
\usepackage[subfigure]{graphfig}

\AtBeginDocument{%
  \providecommand\BibTeX{{%
    \normalfont B\kern-0.5em{\scshape i\kern-0.25em b}\kern-0.8em\TeX}}}

\setcopyright{acmcopyright}
\copyrightyear{2018}
\acmYear{2018}
\acmDOI{10.1145/1122445.1122456}

\acmJournal{JACM}
\acmVolume{37}
\acmNumber{4}
\acmArticle{111}
\acmMonth{8}

\begin{document}

\title{Survey on Deep Multi-modal Data Analytics: Collaboration, Rivalry and Fusion}

\author{Yang Wang}
\email{yangwang@hfut.edu.cn}
\affiliation{%
  \institution{Key Laboratory
of Knowledge Engineering with Big Data (Hefei University
of Technology), Ministry of Education; the School of Computer Science
and Information Engineering, Hefei University of Technology,
China.}
}

%
%
%
%

\begin{abstract}
  With the development of web technology, multi-modal or multi-view data has surged as a major stream for big data, where each modal/view encodes individual property of data objects. Often, different modalities are complementary to each other. Such fact motivated a lot of research attention on fusing the multi-modal feature spaces to comprehensively characterize the data objects. Most of the existing state-of-the-art focused on how to fuse the energy or information from multi-modal spaces to deliver a superior performance over their counterparts with single modal. Recently, deep neural networks have exhibited as a powerful architecture to well capture the nonlinear distribution of high-dimensional multimedia data, so naturally does for multi-modal data. Substantial empirical studies are carried out to demonstrate its advantages that are benefited from deep multi-modal methods, which can essentially deepen the fusion from multi-modal deep feature spaces. In this paper, we provide a substantial overview of the existing state-of-the-arts on the filed of multi-modal data analytics from shallow to deep spaces. Throughout this survey, we further indicate that the critical components for this field go to collaboration, adversarial competition and fusion over multi-modal spaces. Finally, we share our viewpoints regarding some future directions on this field.
\end{abstract}


\maketitle

\section{Introduction}
Multi-modal/view\footnote{modal and view are alternatively used throughout the rest of the paper.} data objects have attracted substantial research attention due to its heterogeneous feature spaces with each of them corresponding to an distinct property, while they are complementary to each other. To well describe multi-modal data objects, the existing research focused on how to leverage the information from multi-modal feature spaces, so as to outperform their single view counterparts. To this end, a lot of research towards leveraging the information from multi-modal spaces have been proposed \cite{WangTIP2015,IJCAI16,WangTNN2017,WangTNN2018,YangIVC2017,YangMM14,YangTIP17,YangNN18,YangMM15,YangSIGIR,YangKAIS16,YangMM13,YangTIP19} , which have been applied to data clustering \cite{IJCAI16,RMVSC,NIPS11,zhang2018generalized,Wang2020TMM,YangInfo13,M-V-C,CVPR12,binarymultiview}, classification \cite{li2018probabilistic,qi2016volumetric}, similarity research \cite{YangTIP19,YangSIGIR,YangIVC2017,5yan2020deep,YangyangTIP19} and others \cite{Meng2015} etc. This fact is theoretically validated by \cite{Dacheng14}, where it indicated that more views will provide more informative knowledge so as to promote the performance. Some methods are also proposed for robust or incomplete multi-view models \cite{LiuTPAMI1,LiuTPAMI2}.
According to the fusion strategy over multi-modal information, the above literature can be roughly classified into \emph{early fusion} \cite{CVPR12}, \emph{late fusion} \cite{Meng2012,Meng2009,diversitymulti} and \emph{collaborative fusion} \cite{IJCAI16,WangTNN2017}.

Specifically, \emph{early fusion} fused the multi-view information, which is subsequently processed by the specific models; \emph{late fusion} performed the specific algorithm over each view independently, then combine the output from each view to be the final outcome. Different from them, as claimed by Wang et al \cite{IJCAI16,WangTNN2017}, \emph{collaborative fusion} is more intuitive than both \emph{early fusion} and \emph{late fusion}, since it can effectively promote the collaboration among multi-modals to achieve the ideal \emph{consensus} \cite{feiping2,Feiping1,feiping3,feiping4,feiping5}, which is critical to multi-modal data analytics, and be widely verified by a large range of applications, such as classification, clustering and retrieval etc.

With the development of web technology, big data has become ubiquitous everywhere, along with the vast access to efficient toolkit, such as GPU, which largely pushed forward the reviving of deep neural networks, to well model the nonlinear distributions of high-dimensional data objects and complex relationship, to further extract the high-level feature representations. A surge of related research on deep multi-modal architectures, which consisted of multiple deep networks with each branch corresponding to one modal, have been proposed to further improve the performance of the shallow models. Beyond that, another featured interaction for deep multi-modal data lies in the Generated Adversarial Networks (named \textbf{GAN} for short), which offered a new manner of multi-modal collaboration via a ``rivalry" fashion, but improved each other to better solve the tasks.

In what follows, we revisit the typical state-of-the-art multi-modal methods especially the deep multi-modal architectures.

\section{Deep Multi-view/modal Learning}

It is widely known that deep network has become the powerful toolkit in the field of pattern recognition \cite{YangCV1,YangCV2}.  That largely improved the performance of the traditional multi-modal models, yielding the so called deep multi-view/modal models, which can effectively extract the high-level nonlinear feature representation to benefit the tasks. In this section, we discuss some recent research of deep multi-modal models for fundamental  problems in terms of clustering and classification, which is often of great importance for other applications.

\subsection{Deep Multi-modal methods for Clustering and Classification}

The literature so far demonstrated that multi-modal tasks solved by deep networks have achieved remarkable performance in fundamental problems such as clustering and classification, which is validated by significant progress below.

Specifically, Zhang et al. \cite{zhang2018generalized} proposed a latent multi-view subspace clustering method (LMSC), which can integrate multiple views into a comprehensive latent representation and encode multi-view complementary information. In particular, it utilized a multi-view subspace clustering method based on deep neural networks to explore more general relationships between different views. Huang et al. \cite{huang2019multi} proposed a multi-view spectral clustering network called MvSCN for brevity, which defined the local invariance of views with a deep metric learning network. MvSCN optimized the local invariance of a single view and the consistency of different views, so that it can obtain the ideal clustering results.

Beyond the above, deep multi-modal models contributed a lot to classification . Specifically, Hu et al. \cite{hu2017sharable} proposed multi-view metric learning (MvML) method based on the optimal combination of various metric distance from multiple views for a joint learning. Based on that, the author further extended MvML to the method, named MvDML, by using a deep neural network to construct a nonlinear transformation of the high-dimensional data objects. Li et al. \cite{li2018probabilistic} proposed a probabilistic hierarchical model for multi-view classification. The proposed method projected latent variables fused with multi-view features into multiple observations. Afterwards,  the parameters and the latent variables are estimated by the Expectation-Maximization (EM) method \cite{EMPRML}.

There are also some methods that focused on the object classification of three dimensional data from multiple views. Specifically, Qi et al. \cite{qi2016volumetric} proposed two different deep architectures of volumetric CNNs to classify three dimensional data. Their approach analyzed the performance gap between volumetric CNNs and multi-view CNNs based on some factors such as deep network architecture and three dimensional resolution, and then proposed two novel architectures of volumetric CNNs. Yang et al. \cite{yang2018complex} argued that it was unreasonable to assume that multi-modal labels are consistent in the real world. Based on that, they further proposed a novel multi-mode multi-instance multi-label deep network (M3DN), which learned label prediction and exploited label correlation simultaneously, in accordance with the optimal transmission (OT) theory \cite{YangIJCAI19}. Chen et al. \cite{chen2017deep} proposed a novel deep multi-modal learning method for multi-label classification \cite{YangIJCAI20}, by exploiting the dependency between labels and modalities to constrain the objective function. In addition, they proposed an effective method for training deep learning networks and classifiers to promote multi-modal multi-label classification. Kan et al. \cite{kan2016multi} proposed a multi-view deep network (MvDN) that eliminated the bias between views to achieve cross-view recognition tasks. This idea sought for non-linear discriminants and view-invariant representations between multiple views, which consisted of two parts: a view-specific sub-network and a common sub-network shared by all views.

To ease the understanding, we summarized some deep multi-view learning algorithms for classification in Table. \ref{tab3}.

\begin{table}[htbp]
	\center
	\small
	\caption{Summary of the Deep Multi-view Learning Algorithms for Classification.}
	\begin{tabular}{p{1.5cm}|p{2.5cm}|p{2.5cm}|p{2.5cm}|p{2.5cm}}
		\hline
		Method & Experiment Datasets & Views & Applications & Characteristic\\
		\hline
		MvML \cite{hu2017sharable} & LFW, KinFaceW-II, VIPeR & LBP, HOG,SIFT & Visual recognition & Separate and sharable multiview metric learning   \\
		\hline
		
		MvLDAN \cite{hu2019multi} &  Noisy MNIST , nMSAD, Half MNIST, Pascal VOC, NUS-WIDE & MV1, MV2, SAD, Left, Right, Image, Tag& Cross-view recognition and retrieval   & Eliminating the divergence of various views         \\
		\hline
		
		HMMF \cite{li2018probabilistic} & Synthetic data, Biomedical dataset, Wiki text-image & Color, Texture, Geometry, Image, Text  & Multi-view Classification  & Multi-view fusion  \\
		\hline
		
		MVCNN-MultiRes \cite{qi2016volumetric} & ModelNet, Real-world reconstructions  & - & Object classification on 3D data  & Two novel volumetric CNNs architectures \\
		\hline
		
		M3DN \cite{yang2018complex} & IAPR TC-12, FLICKR25K, MS-COCO, NUS-WIDE WKG Game-Hub & Image, Text & Complex object classification & Multimodal Multi-instance Multi-label (M3) learning \\
		\hline		
		
		MvDN \cite{kan2016multi} & MultiPIE, FRGC, LFW  & Poses & Cross-view face recognition &  Multi-view deep network \\
		\hline
	\end{tabular}
	\label{tab3}
\end{table}

\subsection{Deep Multi-view Learning for Other Multimedia Applications}

Multi-view learning methods have also been widely applied in the area of multimedia analytics, mainly including image retrieval, image representation and image recognition. In this section, we discussed some typical algorithms as follows.

There have been several deep multi-view learning methods \cite{3GuoMulti,5yan2020deep} that are proposed to deal with image retrieval. A multi-view 3D object retrieval approach \cite{3GuoMulti} was proposed based on a deep embedding network, which is jointly supervised by triplet loss and classification loss. After generating a comprehensive description for each 3D object by extracting deep features from different views, the multi-view 3D object retrieval could be formulated as a set-to-set matching problem. A lot of deep multi-modal hashing models \cite{5yan2020deep,YangTIP19} designed feature descriptors into a similarity preserving hamming space to perform large-scale retrieval. For example, Yan et al. \cite{5yan2020deep} exploited a view stability evaluation to obtain the relationship among views. Meanwhile, the above methods integrated multi-modal dependencies that are designed to obtain hash code, which preserve the merits of convolutional network from multi-view perspective.

For deep image representation, multi-view learning methods \cite{4Yang,27ma2017towards} have been proposed to represent the image based on deep auto-encoder. SkeletonNet \cite{4Yang} proposed a hybrid deep network based on deep auto-encoder, which adopted skeleton embedding process for unsupervised multi-view subspace learning. Throughout the tensor factorization, it takes full use of the strength of both shallow and deep architectures when dealing with low-level and high-level views. Bimodal Correlative
Deep Autoencoder (BCDA) \cite{27ma2017towards} constructed fashion semantic space based on the image-scale space. Then it captures the internal correlations between visual features and fashion styles by utilizing the intrinsic matching rules of tops and bottoms.

Beyond the above, deep multi-view learning methods \cite{9ding2015robust,14wu2018coupled,23lin2018contactless,29niu2018multi,30song2018deep} are widely employed in the field of image recognition. Deep Multi-Modal Face Representation (MM-DFR) \cite{9ding2015robust} first adopted multiple Convolutional Neural Networks (CNN) to extract features for face images. Then all these features are concatenated as a raw feature vector. To reduce the dimension the feature vector,  auto-encoder is adopted to compress the raw feature. Coupled Deep Learning (CDL) \cite{14wu2018coupled} transformed the heterogeneous face matching problem as the homogeneous face matching based on deep model. Furthermore, it supplements trace norm and a block-diagonal prior to the loss function, to enhance the relevance between the projection matrices of images from different views. Lin et al. \cite{23lin2018contactless} proposed a 3D fingerprint segmentation network and a multi-view generation network to obtain the deep representation to better match 3D fingerprint. The recognition performance could be improved with the global shape and texture information by integrating respective 3D contour map. Niu et al. \cite{29niu2018multi} proposed a multi-scale CNN architecture to extract deep features at multiple scales corresponding to visual concepts of different levels. The proposed architecture solved the problem of variable label number by explicitly estimating the optimum label quantity. Moreover, multi-modal extension is formulated to utilize the noisy user-provided tags.

Multi-Modal CNN for Multi-Instance Multi-Label (MMCNN-MIML) \cite{30song2018deep} first generated multi-modal instances from the image and text description. Then, it concatenated the visual instances and its corresponding context embedding, which directly incorporated the groupings of class labels into the network by sharing the layers among related labels within the same group. Throughout this way, it performed the final predictions on the bag level to solve multi-label classification.

To study the problem of multi-view learning via an end-to-end network, Jia et al. \cite{8jia2019deep} took multi-view learning criteria into consideration, while proposed a neuron-wise correlation-maximizing regularizer, named CorrReg. Particularly, CorrReg belonged to an intermediate network layer to concatenate the features of individual views, leading to the superior classification performance. Deep Explicit Attentive Multi-View Learning Model (DEAML) \cite{12gao2019explainable} is a deep multi-view learning method which focused on the task of recommender system. DEAML first adopted an explainable deep structure to construct an initial network. Meanwhile, it developed an attentive multi-view learning framework to optimize key variables with an explainable structure, which can largely improve the recommendation precision. Zhang et al. \cite{20zhang2019cpm} aimed to deal with the problem of incomplete view and constructed a novel framework, namely Cross Partial Multi-View Networks (CPM-Nets). The encoding networks in CPM-Nets ensured that all samples and all views can be jointly exploited regardless of view-missing patterns. Furthermore, it jointly considered multi-view complementary information and class distribution, which enforced all views to learn from each other and enhanced the classification performance.

Deep multi-view learning methods are also well applied to other applications. For taxi demand prediction, Yao et al. \cite{22yao2018deep} proposed a Deep Multi-View Spatial-Temporal Network (DMVST-Net) framework which considered three views simultaneously, including temporal view, spatial view and semantic view, respectively. Extensive experiments on a large-scale taxi request dataset from ``Didi Chuxing" validated the performance of DMVST-Net. Multi-view Subspace Learning to Rank (MvSL2R) aimed to deal with the problem of learning to rank from multiple information sources. MvSL2R introduced an autoencoder to capture the information of feature mappings from both intra-modal and cross-modals. Furthermore, an end-to-end architecture is constructed to learn both the joint ranking objective and the individual rankings. Medical concept normalization is also a specific task in biomedical research. In order to address non-standard expressions and short-text problem, Luo et al. \cite{26luo2018multi} proposed a multi-view CNN to extract semantic matching patterns and learned to synthesize them from different views. Meanwhile, multi-task shared structure is also adopted to obtain medical correlations between disease and procedure, which is beneficial for the disambiguation tasks.

Wang et al. \cite{10wang2015learning} proposed a deep multi-modal method to learn compact hash codes, which can fully exploit intra-modality and inter-modality correlations to learn accurate representations. Furthermore, an orthogonal regularizer is proposed to address the challenge of redundancy lying in the binary codes. Yu et al. \cite{15yu2016deep} developed a novel deep multi-modal distance metric learning (named \textbf{Deep-MDML}) method, which well retrieved the images given the queries. Deep-MDML learned a nonlinear distance metric instead of a linear one. It adopted gradient descent optimization to assign optimal weights and distance metric for different modalities. Furthermore, local geometry (encoding the
visual similarity) is combined with the structure regression (encoding the penalty of click features) to obtain a novel objective function which benefited from both the click features and visual features.

As aforementioned, one critical issue for multi-modal learning is to promote the effective multi-modal collaborations for multi-modal information fusion. Unlike the research stated in previous sections, another adversarial correlations among multi-modal is proposed, which is motivated by Generative Adversarial Networks (GANs). In what follows, we revisit the preliminaries, followed by the research on its deep multi-modal applications.

\subsection{Generative Adversarial Networks and Its Applications}

The traditional generative adversarial networks (GANs) \cite{r1} consist of two parts: a \emph{generator} and a \emph{discriminator}. The generator is able to generate realistic samples to confuse the discriminator. Such objective is formulated by a minimax game between a generator and a discriminator, which compete with each other to synthesize more realistic samples and identify the real samples. Mathematically, the GANs framework is to optimize the following:

\begin{equation}
\label{eq1}
\min\limits_{\theta}\max\limits_{\phi}V\left(D,G\right)=E_{x\thicksim P_{data}\left(x\right)}\left[\log\left(D\left(x\right)\right)\right]+E_{z\thicksim P\left(z\right)}\left[\log\left(1-D\left(G\left(z\right)\right)\right)\right],
\end{equation}
where
\begin{itemize}
\item $G\left(\bullet\right)$ denotes the generator parameterized by $\theta$,
\item $D\left(\bullet\right)$ denotes the generator parameterized by $\phi$.
\item $P_{data}\left(x\right)$ denotes the raw data distribution, and
\item $P\left(z\right)$ denotes a prior distribution.
\end{itemize}
By optimizing the above objective function, the generator is expected to generate realistic samples with real ones. The framework on general training process is shown in Algorithm \ref{algorithm1}.

\begin{algorithm}[htb]
\caption{General training process of generative adversarial networks}
\label{algorithm1}
\begin{algorithmic}[1]
	\Require
 Initialize $\theta$ for $G\left(\bullet\right)$ and $\phi$ for $D\left(\bullet\right)$\\
 \textbf{for} number of training iterations \textbf{do}\\

 \qquad \textbf{for} $k$ steps \textbf{do}\\

 \qquad \quad Sample $m$ real samples $\{x^1,x^2,\dots,x^m \}$ from the raw data distribution $P_{data}\left(x\right)$\\

 \qquad \quad Sample $m$  noise samples  $\{z^1,z^2,\dots,z^m \}$   from the prior distribution  $P\left(z\right)$\\

 \qquad \quad Given the generator $G\left(\bullet\right)$ \\

 \qquad \quad Update the discriminator parameters $\phi$  to maximize\\

 \qquad \quad $V= \frac{1}{m} \sum^m_{i=1}\left[\log{\left(D\left(x^i\right)\right)}+\log{\left(1-D\left(G\left(z^i\right)\right)\right)}\right]$\\

 \qquad \quad $\phi \leftarrow \phi + \eta \nabla V\left(\phi\right)$ \\

 \qquad \textbf{end for}\\

  \qquad Sample another $m$ noise samples $\{z^1,z^2,\dots,z^m \}$  from the prior distribution  $P\left(z\right)$\\

 \qquad Update the generator parameters $\theta$  to minimize\\

 \qquad $V=\frac{1}{m}\sum^m_{i=1}\log{\left(1-D\left(G\left(z^i\right)\right)\right)}$\\

 \qquad $\theta \leftarrow \phi + \theta \nabla V\left(\theta\right)$ \\

 \textbf{end for}

\end{algorithmic}
\end{algorithm}

Several studies further attempted to improve the quality of generated samples. Denton et al. \cite{r2} proposed a generative parametric model capable of generating high quality samples of natural images, which adopted a cascaded of convolutional networks within a Laplacian pyramid framework to generate samples. Radford et al. \cite{r3} introduced a class of CNNs called DCGANs, by making up strided convolution layers and batch normalization to the network. Li et al. \cite{r4} proposed an encoder-and-decoder architecture so that it can generate better results, and further modified the basics on a conditional generative adversarial network to generate realistic clear images. Xu et al. \cite{r5} proposed an attentional generative adversarial network (AttnGAN), which can synthesize fine-grained details at different subregions of the image by paying attentions to the relevant words in the natural language description. Mirza et al. \cite{r6} proposed a conditional version of generative adversarial nets, and a generator conditioned on the class labels greatly improved the quality of the generated samples.

In addition, the traditional GAN has been adopted in supervised, semi-supervised and unsupervised learning methods for classification. Baumgartner et al. \cite{r7} developed a novel feature attribution technique based on wasserstein generative adversarial networks (WGANs) in supervised classification, the method could obtain visual attribution and classification results. Choi et al. \cite{r8} proposed StarGAN for multi-view learning, which mainly focused on cross-domain data generation to obtain better visual attribution. Deng et al. \cite{r10} proposed structured generative adversarial networks (SGANs) for semi-supervised image classification. SGAN could generate the images with high visual quality and strictly follow the designated semantic. Salimans et al. \cite{r11} presented a variety of new architectural features and training procedures to apply to the GANs framework, leading to the ideal results in semi-supervised classification. Furthermore, both ALI \cite{r12} and Triple-GAN \cite{r13} learned a generator network and an inference network throughout an adversarial process, where they are specifically designed for semi-supervised classification.

Chen et al. \cite{r14} proposed the method, named InfoGAN, by maximizing the mutual information between a small subset of the latent variables and the observation, the InfoGAN is effective to learn disentangled representations in a completely unsupervised manner. Dizaji et al. \cite{r15} proposed a novel semi-supervised deep generative model to target gene expression inference, based on GAN to approximate the joint distribution of landmark and target genes. More accurate prediction results were obtained for the types of gene expression.

\subsection{Deep Generative Adversarial Networks for multi-modal clustering}

GAN based clustering models attract great attention. Specifically, Dizaji et al. \cite{r16} proposed a deep unsupervised hashing function, called HashGAN, which efficiently obtained the binary representation of the input images to achieve sufficiently good performance in single-view image clustering. Zhou et al.\cite{r17} developed a novel unsupervised deep subspace clustering model following a GAN framework, which was termed deep adversarial subspace clustering (DASC) that learned to supervise the generator by evaluating clustering quality via an unsupervised manner. Mukherjee et al. \cite{r18} proposed ClusterGAN as a mechanism for clustering using GANs. By sampling latent variables from a mixture of one-hot encoded variables and continuous latent variables, coupled with an inverse network.   Dizaji et al. \cite{r19} proposed a generative adversarial clustering network, called ClusterGAN, which adopted the adversarial game in GAN for the clustering task in single-view samples, and employed an efficient self-paced learning algorithm to boost its performance.

\begin{figure}
\centering
\includegraphics[width=.8\textwidth]{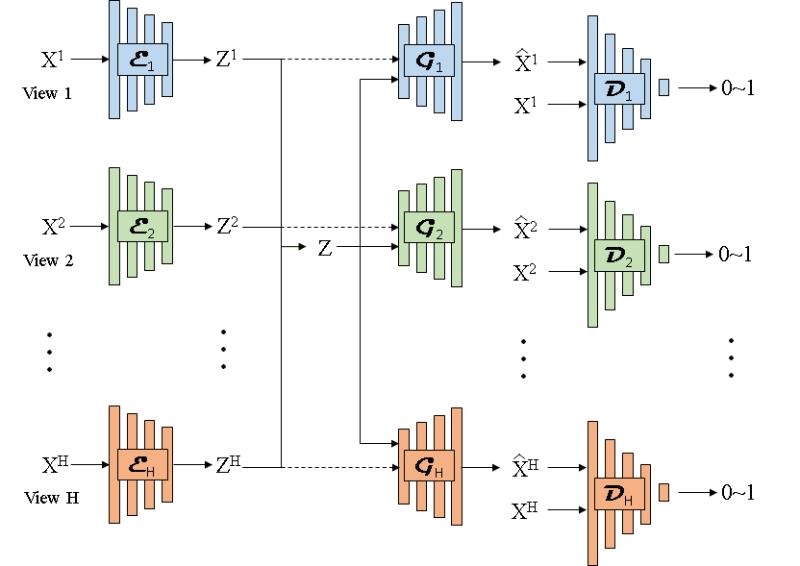}
\caption{The basic deep architecture of generative adversarial network based models for multi-view clustering. For the i-th view $X_i$, the encoder, denoted as $E_i$, yielded to deep representation, denoted as $Z^{i}$, that served as the input to generator $G_i$, where it simultaneously served as decoder (paired with encoder) and generator itself $\widehat{X_i}$ with ground-truth $X_i$, which competed to fool discriminator $D_i$, to play the min-max rivalry game. All the above are jointly trained, then combine $Z^{i}$ to be $Z$ as multi-view representation for final clustering.}
\label{fig:multigan}
\end{figure}

Upon the traditional model, a lot of work have been proposed in the field of Multi-modal GAN based models. Despite the variant of the existing work, the basic deep architecture is illustrated in Fig.\ref{fig:multigan}, which motivates substantial research output.

In particular, Wang et al. \cite{r20} proposed Consistent GAN for the two-view problem, which utilized one view to generate the missing data of the other view, and then performed clustering on the generated complete data. Yang et al. \cite{r21} proposed deep clustering via a Gaussian mixture variational auto-encoder (VAE) with graph embedding, taking Gaussian mixture model (GMM) as the prior to facilitate clustering in VAE and applied graph embedding to handle data with complex spread. Tao et al. \cite{r22} proposed an adversarial graph auto-encoders (AGAE) model to incorporate ensemble clustering into a deep graph embedding process, and developed an adversarial regularizer to guide the network training with an adaptive partition-dependent prior. Sage et al. \cite{r23} proposed to adopt the synthetic labels obtained throughout clustering to disentangle and stabilize GAN training, and validated this approach on single-view images to demonstrate its generality. Jiang et al. \cite{r24} proposed a deep adversarial learning framework for clustering with mixed-modal data. In particular,the framework learned the mappings across individual modality spaces by virtue of cycle-consistency. Li et al. \cite{r25} constructed a novel deep adversarial multi-view clustering (DAMC) framework, which developed multi-view auto-encoder networks with shared weights to learn effective mapping from original features to a common low-dimensional embedding space. Xu et al. \cite{r26} presented an adversarial incomplete multi-view clustering (AIMC) method, which sought the common latent space of multi-view data and performed missing data inference simultaneously.

\begin{table}[htbp]
	\center
	\small
	\caption{Summary of Some Studies in the Application of Generative Adversarial Networks: Task Type (Classification, Clustering), Dataset Used, View (Single-view, Multi-view) and a Brief Method Description}
	\begin{tabular}{p{1.7cm}|p{1.7cm}|p{0.65cm}|p{7.0cm}}
		\hline
		Method & Datasets & Views & Method \\
		\hline
		Classification \cite{r11} & CIFAR-10, MNIST, SVHN, ImageNet & S & Presented a variety of new structural features and training procedures based on the GAN framework. The method focused on two applications of GANs: semi-supervised learning, and the generation of images to be visually realistic.\\
		\hline
		Classification \cite{r8} &CelebA, RaFD & S & Proposed a scalable image-to-image translation model among multiple domains using a single generator and a discriminator. StarGAN generated images of higher visual quality compared with the existing methods.\\
		\hline
		Classification \cite{r10} &CIFAR-10, MNIST, SVHN & S & Proposed SGAN for semi-supervised conditional generative modeling, which learned from a small set of labeled instances to disentangle the semantics of our interest from other elements in the latent space.\\
		\hline
		Classification \cite{r12} &CIFAR-10, SVHN, CelebA& S &proposed the adversarially learned inference (ALI) model, which jointly learned a generation network and an inference network using an adversarial process.\\
		\hline
		Classification \cite{r13} &CIFAR-10, MNIST& S &Consisted of a unified game-theoretical framework with three players-a generator, a discriminator and a classifier, to conduct semi-supervised learning with compatible utilities.\\
		\hline
		Clustering \cite{r26} &Reuters, BDGP, Youtube& M &Tried to seek a common high-level representation for incomplete multi-view data. Moreover, the hidden information of the missing data is captured by the missing data inference via the element-wise reconstruction and the GAN.\\
		\hline
		Clustering \cite{r24} &NUS-WIDE-10K, WikipediaT& M &Unified the modality-specific representations by learning the cycle-consistent mappings across modalities via an adversarial manner. Subsequently, the method performed a common clustering with the unified representations.\\
		\hline
		Clustering \cite{r18} &MNIST, Fashion-10& S &Proposed ClusterGAN, an architecture that enabled clustering in the latent space, which considered discrete-continuous mixtures for sampling noise variables.\\
		\hline
		Clustering \cite{r20} &HW, BDGP, MNIST& M &Simultaneously learned an excellent clustering structure and inferred the incomplete views on the common subspace structure via GAN model.\\
		\hline
		Clustering \cite{r19} &CIFAR-10, MNIST& S &Defined a minimum entropy loss on the real data along with a balanced self-paced learning algorithm based on the GAN to enhance the training for clustering.\\
		\hline
		Clustering \cite{r16} &CIFAR-10, MNIST& S &deployed the tied discriminator and encoder, along with the adversarial loss as a data-dependent regularization for unsupervised learning of hash function.\\	
		
		\hline
	\end{tabular}
	\label{tab1}
\end{table}

\subsection{Datasets for deep multi-modal GAN-based clustering methods}
A number of benchmark datasets for GAN-based clustering methods in multi-view samples have been proposed, which consist of Reuters, HW, BDGP, MNIST, CCV, and Youtube. Important statistics are summarized in Table 2 and a brief descriptions of the datasets is presented below. Reuters \cite{r30} consists of 111740 documents written in 5 languages of 6 categories represented as TFIDF vectors. A subset of the Reuters database consists of English, French and German as three views. For each category, 500 documents are randomly chosen. Totally 3000 documents are used. Handwritten numerals (HW) \cite{r27} database consists of 2,000 images for 10 classes from 0 to 9 digit. Each class contains 200 samples. Each sample has 2 kinds of features: 76 Fourier coefficients and 216 profile correlations. BDGP \cite{r28} is a two-view database. One is visual view and the other is textual view. It contains 2,500 images about drosophila embryos belonging to 5 categories. Each image is represented by a 1,750-D visual vector and a 79-D textual feature vector. MNIST \cite{r29} is a handwritten digits image database with the resolution as 28 $\times$ 28. MNIST consists of 60,000 training examples and 10,000 testing examples. The first view is the original dataset, and the second view is constructed based on the first view with additive noise. The Columbia Consumer Video (CCV) dataset \cite{r31} contains 9,317 YouTube videos with 20 diverse semantic categories. The subset (6773 videos) of CCV provided by \cite{r31} is used, along with three hand-crafted features: STIP features with 5,000 dimensional bag-of-words (BoWs) representation, SIFT features extracted every two seconds with 5,000 dimensional BoWs representation, and MFCC features with 4,000 dimensional BoWs representation. Youtube \cite{r32} contains 92457 instances from 31 categories, each described by 13 feature types. For example, 500 instances from each category are sampled. 512-D vision feature, 2000-D audio feature and 1000-D text feature as three views are selected.
\begin{table}[htbp]
	\center
	\small
	\caption{Summary of the Features of the Datasets for GAN-based clustering methods in multi-view samples}
	\begin{tabular}{p{1.2cm}|p{2.4cm}|p{7.5cm}}
		\hline
		 Datasets & Features & Multi-view description \\
		\hline
		 Reuters \cite{r30} & 6 categories and 111,740 documents & Three views: English, French and German. The subset consists of 3000 documents from 6 categories.	 \\
		\hline
		
		HW \cite{r27} &10 classes and 2,000 images & Two views: 76 Fourier coefficients and 216 profile correlations.\\
		\hline
		
		BDGP \cite{r28} &5 categories and 2,500 images & Two views: Visual view and textual view. Each image is represented by a 1,750-D visual feature vector and a 79-D textual feature vector.\\
		\hline
		 MNIST \cite{r29} &10 classes and 7,000 images &Two views: The first view is the original dataset, and the second view is the images of the first view with additive noise.\\
		\hline
		
		CCV \cite{r31} &20 categories and 9,000 videos &Three views: STIP features with 5,000-D BoWs representation, SIFT features extracted every two seconds with 5,000-D BoWs representation, and MFCC features with 4,000-D BoWs representation.\\
		\hline
		
		Youtube \cite{r32} &31 categories and 92,457 videos &Three views: 512-D vision feature, 2000-D audio feature and 1000-D text feature.\\
		\hline
	\end{tabular}
	\label{tab2}
\end{table}

\subsection{Deep Generative Adversarial Networks for Multi-View Applications}
Other multi-view applications, such as cross-view joint distribution matching, multi-view action recognition, multi-view facial expression recognition and face image synthesis, multi-view networks analysis etc., are addressed effectively via adversarial learning. Du et al~\cite{DBLP:conf/kdd/DuDXZW18} constructed a multi-view adversarially learned inference model, named \textbf{MALI}, to achieve cross-domain joint distribution matching. Based on shared latent representations of two different views, \textbf{MALI} generated arbitrary number of paired synthetic samples to support mapping learning. For multi-view human action recognition, Wang et al.~\cite{DBLP:conf/iccv/WangDTLF19} proposed a generative multi-view action recognition method, named \textbf{GM-VAR}, to bridge the gap between different feature domains. By exploring the latent connection knowledge in both intra-view and cross-view, the proposed method generated one latent representations for one view conditioning on the other view. Besides, the high-level label information is incorporated by a view correlation discovery network. Xuan et al.~\cite{DBLP:journals/tie/XuanCLHBZ19} employed multi-view generative adversarial network to address automatic pearl classification problem. Particularly, they developed a multi-view GAN to automatically expand the labeled training set. For multi-view facial expression recognition task, Lai et al.~\cite{DBLP:conf/fgr/LaiL18} proposed a GAN based multi-task learning approach to learn the emotion-preserving representations. This method utilizes generator to frontalize input non-frontal face images into frontal face images while preserving the identity and expression characteristics. Cao et al.~\cite{DBLP:journals/corr/abs-1802-07447} presented Load Balanced Generative Adversarial Networks (LB-GAN) to address multi-view face imagine synthesis, which decomposes the synthesis task into two well constrained subtasks supported by two pairs of GAN: face normalizer and face editor. Tian et al.~\cite{DBLP:conf/ijcai/TianPZZM18} studied a multi-view generation method named CR-GAN, which is a two-pathway learning model leveraging labeled and unlabeled data for self-supervised learning to improve generation quality. Sun et al.~\cite{DBLP:conf/ijcai/SunWHTH19} studied the multi-view network representation problem and constructed a GAN based model MEGAN to generate multi-view network embedding, which preserves the knowledge from the individual network views and the correlations across different views. For Learning to Rank problem, Cao et al.~\cite{DBLP:journals/corr/abs-1801-10402} proposed to capture information from different views. They introduced a generic framework for multi-view subspace learning to rank, named MvSL2R, based on which two multi-view learning solutions are developed.

One of the obstacles in multi-view applications is the problem of missing view or missing data. Recently, several studies are proposed to tackle this challenge. For example, Chen et al.~\cite{DBLP:journals/corr/ChenD16a} proposed Multi-view BiGAN, a BiGAN based model that can deal with missing views. Doinychko et al.~\cite{DBLP:conf/ecir/DoinychkoA20} proposed a biconditional GAN model that contained two generators and a common discriminator, which can deal with missing views. By conducting a tripartite game with two generators and a discriminator, this technique can learn missing view conditionally. Shang et al.~\cite{DBLP:conf/bigdataconf/ShangPSCLB17} constructed a GAN based model, termed as VIGAN, to combat missing view imputation problem. This approach handled each view as a separate domain and identified domain-to-domain mappings by adversarial learning, then exploited a multi-modal denoising autoencoder (DAE) to reconstruct the missing view. Kanojia et al.~\cite{DBLP:conf/ncc/KanojiaR20} proposed a conditional generative adversarial networks based model to synthesize an image with the completed missing regions by using multi-view images. To solve multi-view frame missing problem, Mahmud et al.\cite{DBLP:journals/corr/abs-1809-10352} employed conditional generative adversarial network to generate the joint spatial-temporal representation of the missing frame. In this approach, all of the multi-view frames are fused together following a weighted average strategy: synthetic representations within the camera are given more weight when they are close to the missing frame, and representations from other overlapping cameras are given more weight when the available intra-camera frames are far apart.
\section{Deep Cross-Modal Learning}
\begin{figure}
	\centering
	\includegraphics[width=1\textwidth]{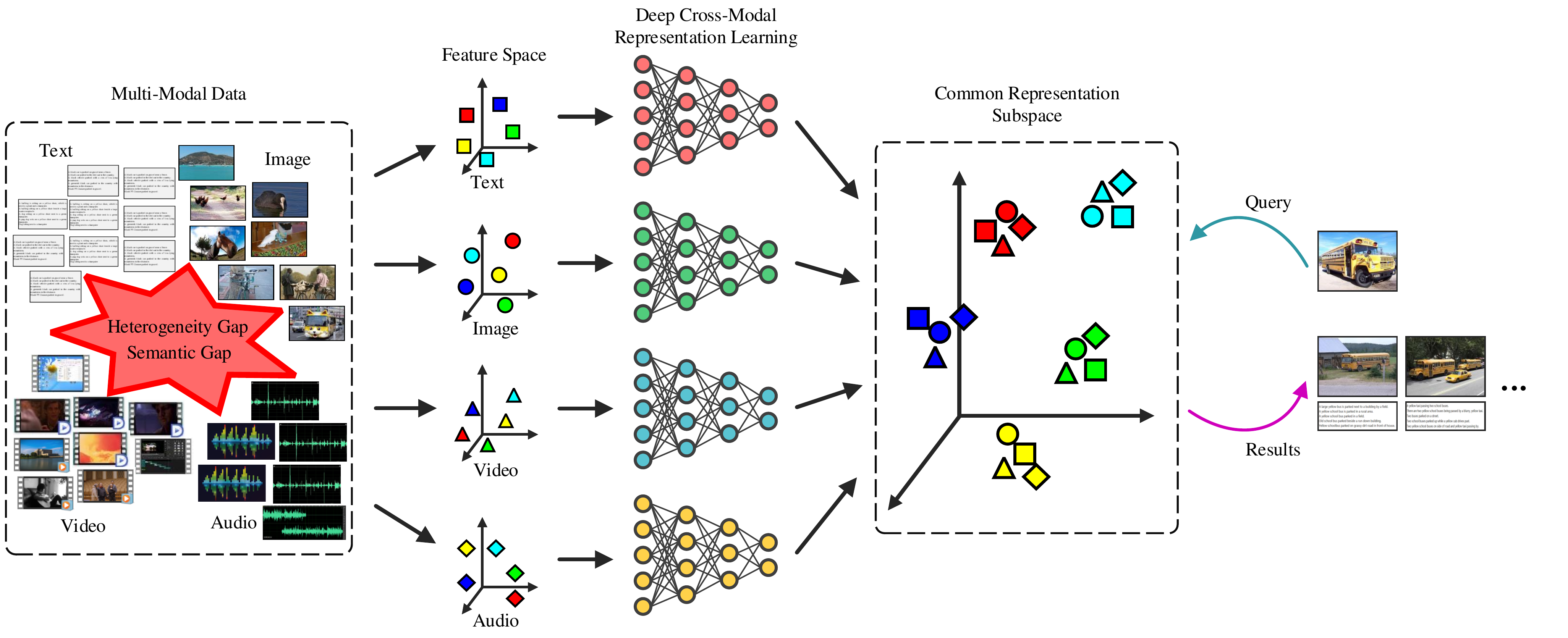}
	\caption{Illustration of deep cross-modal representation learning framework. Text, Image, video and audio represented different modalities, each of which learned a deep neural network for nonlinear feature representation. Based on that, a common subspace is learned, where all the feature representations are projected. The final cross-modal retrieval is conducted against such common representations.}
	\label{fig:deep-crosds-modal}
\end{figure}

Recently, deep cross-modal learning has been broadly discussed and explored. Compared with the cross-modal method that only handled the hand-craft features, the deep cross-modal learning using has become a hot topic of current research, where the basic architecture is shown in Fig.~\ref{fig:deep-crosds-modal}. To evaluate the existing research. There are many benchmark cross-modal datasets, which are summarized in Table \ref{tab4}.

One big question is how to effectively bridge the semantic gap and capture the correlation between heterogeneous modalities. To this end, abundant research has been proposed. In what follows, we discuss several categories for this filed.

\begin{center}
	\renewcommand{\arraystretch}{1.3}
	\centering
	\begin{longtable}{p{3.8cm}|p{1.1cm}|p{1.0cm}|p{2.1cm}|p{3.5cm}}
	\caption{Summary of typical benchmark cross-modal datasets}
	\label{tab4}\\
		\hline
		Datasets & Samples & Classes & Types & Modals \\
		\hline
		NUS-WIDE \cite{chua2009nus}  & 269,648 & 81 & Web images &  Image-text  \\
		
		MIRFlickr-25k \cite{huiskes2008mir} & 25,015  & 24 &   Images   &  Image-text \\
		
		MS COCO \cite{Lin2014Microsoft}  & 123,287 & 80 & Images & Image-text \\
		
		UND Collection X1 \cite{Xin2005IR}  & 4,584 & 82 & Facial corpus & Visible, thermal images\\		 		
		
		Carl \cite{Virginia2013A} & 4,920 & 41 & Images & Visible, Near-infrared, Thermal spectrum   \\
		
		NVESD \cite{Hu2015Thermal} & 900 & 50 & Images & Thermal images, Visible images  \\
		
		Wikipedia \cite{Pereira2014On} & 2,866 & 10 &  Image-text pairs &  Image-text  \\
		
		Pascal Sentence  \cite{Rashtchian2010Collecting} & 1,000 & 20 & Images &  Image-text   \\
		
		VIPeR \cite{Forsyth2008} & 1,264 & 632 & Pedestrian images  & Two different camera views  \\
		
		CUHK01 \cite{Wei2014DeepReID} & 1,942 & 971 & Pedestrian images &  Two cameras  \\
		
		CUHK-PEDES \cite{Li2017Person} & 40,206 & 13,003 & Pedestrian images &  Two cameras  \\
		
		Flickr30K \cite{Young2014From} & 31,783 & - & Images &  Image-text  \\
		
		Caltech-UCSD Birds (CUB) \cite{Reed2016Learning} & 11,788 & 200 & Bird images &  Two cameras   \\
		
		IAPRTC-12 \cite{DBLP:journals/cviu/EscalanteHGLMMSPG10} &20,000 & 102 & image-text pairs & Image-text  \\
		
		SUN RGB-D\cite{Song2015SUN} & 10,355 & 19 & Pedestrian images &   RGB images, Depth images   \\
		
		SUN RGB-D\cite{Silberman2012Indoor} & 1,449 & 10 & Images &  Two cameras  \\
		
		\hline
		
	\end{longtable}
\end{center}

\subsection{Deep Cross-Modal Retrieval}
The existing deep cross-modal retrieval methods can be divided into two categories: (1) real-valued representation learning and (2) binary code or hashing learning. The former aimed to learn a common real-valued subspace, so that the semantic similarity between each pair of cross-modal representations can be measured. The latter aimed to learn quantitative representations in cross-modal Hamming space. Thus the similarity of different modal representations can be measured by Hamming distance. Compared with real-valued representation learning methods, the hashing learning methods enjoyed the higher efficiency in terms of searching and storage.

\subsubsection{Cross-Modal Real-Valued Representation Learning}
There are a large number of research for this category. Specifically,
Andrew et al.~\cite{DBLP:conf/icml/AndrewABL13} extended the traditional subspace learning method via canonical-correlation analysis, named CCA, by deep neural networks. A novel model, named Deep-CCA, is equipped with a two-branches deep neural networks (DNNs), which can learn complex nonlinear transformations of two modalities. Feng et al.~\cite{DBLP:conf/mm/FengWL14} proposed to employ two uni-modal autoencoders, called Corr-AE, to learn cross-modal hidden representations. In this scheme, a novel optimal objective is constructed to minimize a linear combination of intra-modal representation learning errors and inter-modal correlation learning errors. Wang et al.~\cite{DBLP:conf/ictai/WangYM15} proposed to explore highly non-linear semantic correlation by deep-CNN. They developed a regularized deep neural network (RE-CNN) to learn cross-modal semantic mapping. Wei et al.~\cite{DBLP:journals/tcyb/WeiZLWLZY17} proposed to use deep convolutional visual features to address cross-modal retrieval. They proposed a new baseline, named Deep-SM, which employed CNN and (Linear Discriminant Analysis) LDA to learn visual and textual features, to improve the retrieval performance effectively. Peng et al.~\cite{DBLP:journals/corr/PengQHY17} discussed a cross-modal correlation learning model, termed as CCL, to fuse multi-grain features by hierarchical network structure. Different from the above traditional method, they considered intra-modality and inter-modality correlation simultaneously during the feature learning, and balanced the intra-modality semantic category constraints and inter-modality pairwise similarity constraints in the common representation learning. Based on that, they further proposed the cross-media multiple deep network (CMDN) ~\cite{DBLP:conf/ijcai/PengHQ16} to learn cross-modal correlations by hierarchically combining the inter-media and intra-media representations. To explore more semantic information in texts, Yu et al.~\cite{DBLP:conf/pcm/YuLQZLTG18} proposed to utilize graph representations to model the semantic relations between words, which is implemented by a dual-path structure model called GIN. Gu et al.~\cite{DBLP:conf/cvpr/GuCJN018} attempted to integrate two generative models with conventional feature embedding method to capture comprehensive semantic information. To this end, a model called GXN is proposed to learn both the high-level abstract representation and the local grounded representation. Shang et al.~\cite{DBLP:journals/ijon/ShangZZS19} constructed a Dictionary Learning based Adversarial Cross-Modal Retrieval (DLA-CMR). Technically, discriminative features are reconstructed by dictionary learning, and statistical characteristics of each modality is exploited by adversarial learning.

We summarized the results in Table~\ref{tab:results-real-value}, where the comparison results of the above approaches in mAP on three most commonly used datasets: Wikipedia~\cite{Pereira2014On}, NUS-WIDE~\cite{chua2009nus} and Pascal Sentence~\cite{Rashtchian2010Collecting}. I2T and I2T represent the Image-to-Text task and Text-to-Image task, Ave. means the average mAP. In general, the retrieval precisions of all these methods on Wikipedia datasets are lower than the other two datasets. This is mainly because that the categories in Wikipedia is more virtual in the aspect of semantic, such as "history", "royalty \& nobility", and "art \& architecture". Meanwhile, these categories are correlative to each other in high-level concepts, which makes the features of these samples more confusing. By contrast, the categories in NUS-WIDE and Pascal Sentence are more concrete, i.e., "bicycle", "bus", "dog", "sheep", etc. That means the features of different categories are more distinctive and discriminative. Thus, these approaches that could learn discriminative representations of cross-modal samples are more effective.

\begin{table}[]
	\caption{The retrieval performance (mAP score) of Deep-CCA~\cite{DBLP:conf/icml/AndrewABL13}, Corr-AE~\cite{DBLP:conf/mm/FengWL14}, RE-CNN~\cite{DBLP:conf/ictai/WangYM15}, Deep-SM~\cite{DBLP:journals/tcyb/WeiZLWLZY17}, CCL~\cite{DBLP:journals/corr/PengQHY17}, CMDN~\cite{DBLP:conf/ijcai/PengHQ16}, GIN~\cite{DBLP:conf/pcm/YuLQZLTG18}, GXN~\cite{DBLP:conf/cvpr/GuCJN018}, ACMR~\cite{DBLP:conf/mm/WangYXHS17}, CM-GANs~\cite{DBLP:journals/tomccap/PengQ19}, MHTN~\cite{DBLP:journals/tcyb/HuangPY20}, CMST~\cite{DBLP:conf/icmcs/WenHYL19}, AdvCAE~\cite{DBLP:journals/kbs/WangPHS19} and DLA-CMR~\cite{DBLP:journals/ijon/ShangZZS19} on Wikipedia~\cite{Pereira2014On}, NUS-WIDE~\cite{chua2009nus} and Pascal Sentence~\cite{Rashtchian2010Collecting} for image-to-text (I2T) retrieval, text-to-image (T2I) retrieval and average (Ave.) performance. The top two scores are in boldface.} \label{tab:results-real-value}
    \begin{tabular}{|l|l|l|l|l|l|l|l|l|l|l|}
    \hline
    \multicolumn{2}{|l|}{\multirow{2}{*}{Methods}} & \multicolumn{3}{c|}{Wikipedia~\cite{Pereira2014On}} & \multicolumn{3}{c|}{NUS-WIDE~\cite{chua2009nus}} & \multicolumn{3}{c|}{Pascal Sentence~\cite{Rashtchian2010Collecting}} \\ \cline{3-11}
    \multicolumn{2}{|l|}{}                         & I2T  & T2I  & Ave.  & I2T  & T2I & Ave. & I2T    & T2I    & Ave.    \\ \hline
    \multicolumn{2}{|l|}{Deep-CCA~\cite{DBLP:conf/icml/AndrewABL13}}                 & 0.473    & 0.546    & 0.510    & 0.694    & 0.669   & 0.682   & \textbf{0.578}      & 0.529      & 0.553      \\ \hline
    \multicolumn{2}{|l|}{Corr-AE~\cite{DBLP:conf/mm/FengWL14}}                  & 0.460    & 0.547    & 0.504    & 0.687    & 0.666   & 0.677   & 0.539      & 0.533      & 0.536      \\ \hline
    \multicolumn{2}{|l|}{RE-CNN~\cite{DBLP:conf/ictai/WangYM15}}                   & 0.340    & 0.353    & 0.347    & -        & -       & -       & -          & -          & -          \\ \hline
    \multicolumn{2}{|l|}{Deep-SM~\cite{DBLP:journals/tcyb/WeiZLWLZY17}}                  & 0.480    & 0.552    & 0.516    & 0.695    & 0.641   & 0.668   & 0.563      & 0.552      & 0.558      \\ \hline
    \multicolumn{2}{|l|}{CCL~\cite{DBLP:journals/corr/PengQHY17}}                      & \textbf{0.490}    & \textbf{0.613}    & \textbf{0.551}    & \textbf{0.736}    & \textbf{0.713}   & \textbf{0.724}   & \textbf{0.582}      & \textbf{0.576}      & \textbf{0.579}      \\ \hline
    \multicolumn{2}{|l|}{CMDN~\cite{DBLP:conf/ijcai/PengHQ16}}                     & 0.478    & 0.569    & 0.523    & \textbf{0.710}    & \textbf{0.709}   & \textbf{0.710}   & 0.567      & \textbf{0.567}      & \textbf{0.567}      \\ \hline
    \multicolumn{2}{|l|}{GIN~\cite{DBLP:conf/pcm/YuLQZLTG18}}                      & 0.452    & \textbf{0.767}    & \textbf{0.610}    & 0.524    & 0.542   & 0.533   & 0.317      & 0.452      & 0.384      \\ \hline
	\multicolumn{2}{|l|}{GXN~\cite{DBLP:conf/cvpr/GuCJN018}}                      & 0.492    & 0.553    & 0.523    & 0.678    & 0.668   & 0.673   & 0.535      & 0.541      & 0.538      \\ \hline
	\multicolumn{2}{|l|}{DLA-CMR~\cite{DBLP:journals/ijon/ShangZZS19}}                  & \textbf{0.539}    & 0.453    & 0.496    & -        & -       & -       & 0.498      & 0.546      & 0.522      \\ \hline
    \end{tabular}	
\end{table}

\begin{table}[]
	\caption{The Image-to-Text retrieval performance (mAP score) of DCMH~\cite{DBLP:conf/cvpr/JiangL17}, DCHUC~\cite{DBLP:journals/corr/abs-1907-12490}, CDQ~\cite{cao2017collective}, DSPOH~\cite{DBLP:journals/tnn/JinLLXQT19}, CPAH~\cite{DBLP:journals/tip/XieDLLT20}, CMDVH~\cite{DBLP:conf/iccv/LiongLT017}, CM-DVStH~\cite{DBLP:journals/pami/LiongLDT20}, SSAH~\cite{DBLP:conf/cvpr/LiDL0GT18}, RDCMH~\cite{DBLP:conf/aaai/LiuYD00G19}, MGAH~\cite{DBLP:journals/tmm/ZhangP20}, TDH~\cite{Deng2018Triplet}, UDCMH~\cite{wu2018unsupervised}, DBRC~\cite{DBLP:conf/mm/LiHN17}, UDCH-VLR~\cite{DBLP:journals/ijon/WangZCLG20} on IAPRTC-12~\cite{DBLP:journals/cviu/EscalanteHGLMMSPG10}, MIRFlickr-25k~\cite{huiskes2008mir} and NUS-WIDE~\cite{chua2009nus}. The top two scores are in boldface.} \label{tab:results-i2t-binary-value}
    \begin{tabular}{|l|l|l|l|l|l|l|l|l|l|l|}
    \hline
    \multicolumn{2}{|l|}{\multirow{2}{*}{Methods}} & \multicolumn{3}{c|}{IAPRTC-12~\cite{DBLP:journals/cviu/EscalanteHGLMMSPG10}} & \multicolumn{3}{c|}{MIRFlickr-25k~\cite{huiskes2008mir}} & \multicolumn{3}{c|}{NUS-WIDE~\cite{chua2009nus}} \\ \cline{3-11}
	\multicolumn{2}{|l|}{}                         & 16bits  & 32bits  & 64bits  & 16bits  & 32bits & 64bits & 16bits    & 32bits    & 64bits    \\ \hline
	\multicolumn{2}{|l|}{DCMH~\cite{DBLP:conf/cvpr/JiangL17}}                     & 0.535    & 0.553    & 0.567    & 0.729    & 0.758   & 0.760   & 0.523      & 0.529      & 0.552  \\ \hline
	\multicolumn{2}{|l|}{DCHUC~\cite{DBLP:journals/corr/abs-1907-12490}}                    & \textbf{0.630}    & \textbf{0.695}    & \textbf{0.701}    & \textbf{0.878}    & \textbf{0.882}   & \textbf{0.881}   & 0.750      & 0.771      & \textbf{0.791}      \\ \hline
	\multicolumn{2}{|l|}{CDQ~\cite{cao2017collective}}                      & -        & -        & -        & 0.864    & 0.832   & -       & \textbf{0.850}      & \textbf{0.849}      & -       \\ \hline
	\multicolumn{2}{|l|}{DSPOH~\cite{DBLP:journals/tnn/JinLLXQT19}}                    & -        & -        & -        & 0.832    & 0.840   & -       & 0.695      & 0.711      & -      \\ \hline
	\multicolumn{2}{|l|}{CPAH~\cite{DBLP:journals/tip/XieDLLT20}}                     & \textbf{0.626}    & 0.663    & 0.671    & 0.775    & 0.791   & 0.787   & 0.613      & 0.629      & 0.630      \\ \hline
	\multicolumn{2}{|l|}{CMDVH~\cite{DBLP:conf/iccv/LiongLT017}}                    & 0.376    & 0.373    & 0.376    & 0.611    & 0.626   & 0.598   & 0.373      & 0.414      & 0.425      \\ \hline
	\multicolumn{2}{|l|}{CM-DVStH~\cite{DBLP:journals/pami/LiongLDT20}}                 & 0.619    & \textbf{0.744}    & \textbf{0.773}    & \textbf{0.882}    & \textbf{0.916}   & \textbf{0.928}   & \textbf{0.797}      & \textbf{0.854}      & \textbf{0.862}      \\ \hline
	\multicolumn{2}{|l|}{SSAH~\cite{DBLP:conf/cvpr/LiDL0GT18}}                     & 0.539    & 0.564    & 0.587    & 0.779    & 0.789   & 0.794   & 0.659      & 0.666      & 0.667      \\ \hline
	\multicolumn{2}{|l|}{RDCMH~\cite{DBLP:conf/aaai/LiuYD00G19}}                    & -        & -        & -        & 0.772    & 0.773   & 0.779   & 0.623      & 0.624      & 0.627      \\ \hline
	\multicolumn{2}{|l|}{MGAH~\cite{DBLP:journals/tmm/ZhangP20}}                     & -        & -        & -        & 0.685    & 0.693   & 0.704   & 0.613      & 0.623      & 0.628      \\ \hline
	\multicolumn{2}{|l|}{TDH~\cite{Deng2018Triplet}}                              & -        & -        & -        & 0.711    & 0.723   & 0.729   & 0.639      & 0.663      & 0.675      \\ \hline
	\multicolumn{2}{|l|}{UDCMH~\cite{wu2018unsupervised}}                           & -        & -        & -        & 0.689    & 0.698   & 0.714   & 0.511      & 0.519      & 0.524      \\ \hline
	\multicolumn{2}{|l|}{DBRC~\cite{DBLP:conf/mm/LiHN17}}                           & -        & -        & -        & 0.587    & 0.590   & 0.590   & 0.394      & 0.409      & 0.417      \\ \hline
	\multicolumn{2}{|l|}{UDCH-VLR~\cite{DBLP:journals/ijon/WangZCLG20}}                           & -        & -        & -        & 0.740    & 0.741   & 0.746   & 0.557      & 0.598      & 0.632      \\ \hline
    \end{tabular}	
\end{table}

\begin{table}[]
	\caption{The Text-to-Image retrieval performance (mAP score) of DCMH~\cite{DBLP:conf/cvpr/JiangL17}, DCHUC~\cite{DBLP:journals/corr/abs-1907-12490}, CDQ~\cite{cao2017collective}, DSPOH~\cite{DBLP:journals/tnn/JinLLXQT19}, CPAH~\cite{DBLP:journals/tip/XieDLLT20}, CMDVH~\cite{DBLP:conf/iccv/LiongLT017}, CM-DVStH~\cite{DBLP:journals/pami/LiongLDT20}, SSAH~\cite{DBLP:conf/cvpr/LiDL0GT18}, RDCMH~\cite{DBLP:conf/aaai/LiuYD00G19}, MGAH~\cite{DBLP:journals/tmm/ZhangP20}, TDH~\cite{Deng2018Triplet}, UDCMH~\cite{wu2018unsupervised}, DBRC~\cite{DBLP:conf/mm/LiHN17}, UDCH-VLR~\cite{DBLP:journals/ijon/WangZCLG20} on IAPRTC-12~\cite{DBLP:journals/cviu/EscalanteHGLMMSPG10}, MIRFlickr-25k~\cite{huiskes2008mir} and NUS-WIDE~\cite{chua2009nus}. The top two scores are in boldface.} \label{tab:results-t2i-binary-value}
    \begin{tabular}{|l|l|l|l|l|l|l|l|l|l|l|}
    \hline
    \multicolumn{2}{|l|}{\multirow{2}{*}{Methods}} & \multicolumn{3}{c|}{IAPRTC-12~\cite{DBLP:journals/cviu/EscalanteHGLMMSPG10}} & \multicolumn{3}{c|}{MIRFlickr-25k~\cite{huiskes2008mir}} & \multicolumn{3}{c|}{NUS-WIDE~\cite{chua2009nus}} \\ \cline{3-11}
	\multicolumn{2}{|l|}{}                         & 16bits  & 32bits  & 64bits  & 16bits  & 32bits & 64bits & 16bits    & 32bits    & 64bits    \\ \hline
	\multicolumn{2}{|l|}{DCMH~\cite{DBLP:conf/cvpr/JiangL17}}                     & 0.581    & 0.592    & 0.608    & 0.754    & 0.764   & 0.770   & 0.566      & 0.582      & 0.590  \\ \hline
	\multicolumn{2}{|l|}{DCHUC~\cite{DBLP:journals/corr/abs-1907-12490}}                    & \textbf{0.615}    & \textbf{0.666}    & \textbf{0.693}    & \textbf{0.850}    & \textbf{0.857}   & \textbf{0.854}   & 0.698      & 0.728      & \textbf{0.749}      \\ \hline
	\multicolumn{2}{|l|}{CDQ~\cite{cao2017collective}}                      & -        & -        & -        & \textbf{0.848}    & \textbf{0.850}   & -       & \textbf{0.832}      & \textbf{0.848}      & -       \\ \hline
	\multicolumn{2}{|l|}{DSPOH~\cite{DBLP:journals/tnn/JinLLXQT19}}                    & -        & -        & -        & 0.832    & 0.841   & -       & \textbf{0.713}      & 0.731      & -      \\ \hline
	\multicolumn{2}{|l|}{CPAH~\cite{DBLP:journals/tip/XieDLLT20}}                     & \textbf{0.614}    & 0.649    & 0.661    & 0.777    & 0.787   & 0.789   & 0.649      & 0.669      & 0.668      \\ \hline
	\multicolumn{2}{|l|}{CMDVH~\cite{DBLP:conf/iccv/LiongLT017}}                    & 0.381    & 0.383    & 0.381    & 0.612    & 0.610   & 0.600   & 0.371      & 0.359      & 0.424      \\ \hline
	\multicolumn{2}{|l|}{CM-DVStH~\cite{DBLP:journals/pami/LiongLDT20}}                 & 0.604    & \textbf{0.689}    & \textbf{0.720}    & 0.788    & 0.815   & \textbf{0.825}   & 0.698      & \textbf{0.778}      & \textbf{0.782}      \\ \hline
	\multicolumn{2}{|l|}{SSAH~\cite{DBLP:conf/cvpr/LiDL0GT18}}                     & 0.538    & 0.566    & 0.586    & 0.783    & 0.793   & 0.794   & 0.613      & 0.632      & 0.633      \\ \hline
	\multicolumn{2}{|l|}{RDCMH~\cite{DBLP:conf/aaai/LiuYD00G19}}                    & -        & -        & -        & 0.793    & 0.792   & 0.800   & 0.664      & 0.669      & 0.669      \\ \hline
	\multicolumn{2}{|l|}{MGAH~\cite{DBLP:journals/tmm/ZhangP20}}                     & -        & -        & -        & 0.673    & 0.676   & 0.686   & 0.603      & 0.614      & 0.640      \\ \hline
	\multicolumn{2}{|l|}{TDH~\cite{Deng2018Triplet}}                              & -        & -        & -        & 0.742    & 0.750   & 0.755   & 0.665      & 0.676      & 0.680      \\ \hline
	\multicolumn{2}{|l|}{UDCMH~\cite{wu2018unsupervised}}                           & -        & -        & -        & 0.692    & 0.704   & 0.718   & 0.637      & 0.653      & 0.695      \\ \hline
	\multicolumn{2}{|l|}{DBRC~\cite{DBLP:conf/mm/LiHN17}}                           & -        & -        & -        & 0.588    & 0.596   & 0.596   & 0.425      & 0.429      & 0.438      \\ \hline
	\multicolumn{2}{|l|}{UDCH-VLR~\cite{DBLP:journals/ijon/WangZCLG20}}                           & -        & -        & -        & 0.758    & 0.761   & 0.762   & 0.605      & 0.656      & 0.674      \\ \hline
    \end{tabular}	
\end{table}

\subsubsection{Cross-Modal Hashing Learning}
Unlike the real-valued cross-modal representation learning, There are abundant research on cross-modal hashing methods.

\textbf{\underline{Supervised Model.}} Jiang et al.~\cite{DBLP:conf/cvpr/JiangL17} proposed a novel deep cross-modal hashing model, named DCMH, which consisted of feature learning model and hash-code learning model, to outperform the cross-modal hashing with hand-crafted features. Tu et al.~\cite{DBLP:journals/corr/abs-1907-12490} proposed to jointly learn an end-to-end Deep Cross-Modal Hashing network and unified Hash Codes (DCHUC). It jointly learned unified hash codes for cross-modal instances and a pair of hash functions for queries. Cao et al.~\cite{cao2017collective} proposed collective deep quantization to jointly learn deep representations and the quantizers for both modalities. Jin et al.~\cite{DBLP:journals/tnn/JinLLXQT19} proposed a novel end-to-end ranking-based hashing framework, named deep semantic-preserving ordinal hashing (DSPOH). Instead of relying on binary quantization functions, DSPOH learned hash functions with deep neural networks by exploring the ranking structure of feature dimensions. Xie et al.~\cite{DBLP:journals/tip/XieDLLT20} developed a novel deep hashing approach, termed as Multi-Task Consistency Preserving Adversarial Hashing (CPAH), to capture cross-modal semantic correlation. Two modules, i.e., consistency refined module (CR) and multi-task adversarial learning module (MA), are developed to learn semantic consistency information. Liong et al.~\cite{DBLP:conf/iccv/LiongLT017} developed a cross-modal deep variational hashing (CMDVH). Instead of a single pair of projections, CMDVH characterized a couple of deep neural network to learn non-linear transformations from cross-modal sample pairs. In their another work~\cite{DBLP:journals/pami/LiongLDT20}, a cross-modal deep variational and structural hashing (CM-DVStH) is developed, including a deep fusion network with a structural layer to maximize the correlation between different modalities so as to generate unified binaries. Based on ranking technique, Liu et al.~\cite{DBLP:conf/aaai/LiuYD00G19} proposed a deep cross-modal hashing approach (RDCMH). This method utilized the feature and label information to generate a semi-supervised semantic ranking list. Then, it integrated the semantic ranking information into deep cross-modal hashing to optimize the model. Inspired by adversarial learning, Li et al.~\cite{DBLP:conf/cvpr/LiDL0GT18} constructed a self-supervised adversarial hashing (SSAH) method, which exploited two adversarial networks to maximize the semantic correlation and consistency of the representations between different modalities. Meanwhile, a self-supervised semantic network is integrated to explore high-level semantics. Deng et al. \cite{Deng2018Triplet} proposed a triplet-based deep hashing (TDH) network for cross-modal retrieval, which captured relative semantic correlation of various modalities in the supervised manner via triplet labels. Chen et al. \cite{chen2018dual} proposed a tri-stage deep cross-modal hashing method, which exploited two deep networks to generate hash codes of different modalities.

\underline{\textbf{Unsupervised Model}}. Li et al.~\cite{DBLP:conf/mm/LiHN17} proposed a model named Deep Binary Reconstruction (DBRC) network, which directly learned the binary hashing codes in an unsupervised fashion. Wu et al. \cite{wu2018unsupervised} proposed UDCMH framework for cross-modal hash code generation. This method integrated deep learning and matrix decomposition with a binary latent factor model and then conducted multi-modal data searches in a self-learning manner. Considering the underlying manifold structure across multi-modal data, Zhang et al.~\cite{DBLP:journals/tmm/ZhangP20} proposed an unsupervised multi-pathway generative adversarial hashing modal (MGAH), in which the generative model tried to fit the distribution over the manifold structure, and then learned to distinguish the generated data and the true positive data sampled from the correlation graph to improve the recognition accuracy. Wang et al.~\cite{DBLP:journals/ijon/WangZCLG20} introduced an Unsupervised Deep Cross-modal Hashing with Virtual Label Regression (UDCH-VLR) approach, which is an unified framework that simultaneously integrated deep hash function training, virtual label learning and regression.

\underline{\textbf{Results Analysis}}. We summarized the results in Table~\ref{tab:results-i2t-binary-value} and~\ref{tab:results-t2i-binary-value}, where they reported the comparison results of the above-mentioned deep cross-modal hashing methods on IAPRTC-12~\cite{DBLP:journals/cviu/EscalanteHGLMMSPG10}, MIRFlickr-25k~\cite{huiskes2008mir} and NUS-WIDE~\cite{chua2009nus} benchmarks for Image-to-Text retrieval and Text-to-Image retrieval, respectively. The most common lengths of hashing codes are 16 bits, 32 bits and 64 bits. MIRFlickr-25k dataset has totally 25000 images collected from Flickr, belonging to 24 basic categories, such as "bird", "sky", "tree" etc. These categories are relatively independent on semantics, in which the samples have discriminative features. This may be the major reason that the approaches perform the best on two data sets. For image query, CM-DVStH wins the competition by mAP = 0.882, 0.916 and 0.928, much higher than others. But for text query, CDQ shows competitive performance on MIRFlickr-25k and NUS-WIDE.

\subsubsection{Deep Cross-Modal Learning for Other Applications.}
In addition to the above researches, some deep methods for other multimedia applications have been proposed. To name a few: Sarfraz et al. \cite{Sarfraz2016Deep} proposed a deep neural network to capture the highly non-linear mapping relationship between two modes in which the mapping process kept identity information. Aytar et al. \cite{Aytar2017Cross} proposed a novel cross-modal scene dataset, along with an optimized cross-modal convolutional neural network, which learned a modal-independent shared representation. Zheng et al. \cite{Zheng2018Hetero} proposed a novel hetero-manifold regularization (HMR) method, which defined multiple sub-manifolds to jointly supervise the learning of hash functions and multi-modal representations. They verified the advanced performance of HMR on cross-camera re-id task and cross-age face recognition task. Kim et al. \cite{Kim2016Deep} recently proposed a novel descriptor that has better discriminative power and greater robustness to non-rigid image deformation. Zhang et al. \cite{zhang2018deep} proposed a cross-modal projection matching (CMPM) loss and cross-modal projection classification (CMPC) loss, to optimize data distribution and improve image-text matching results. It can be seen that most of the deep cross-modal hashing methods used paired optimization approaches, which have no advantages for large scale datasets. In \cite{marin2019recipe1m}, a large-scale structured corpus is proposed and a neural network is trained. In order to do so, they used deep neural networks to combine images for retrieval tasks. Li et al. \cite{liu2019mtfh} proposed encoding hashing heterogeneous data with varied lengths, which used an effective objective function to learn the semantic correlation of different modalities. Such hash encoding method is robust to various challenging tasks for cross-modal search. Yuan et al. \cite{yuan2019acm} proposed an adaptive cross-modal (ACM) feature learning framework based on graph convolutional neural network for RGB-D scene recognition. This method adaptively selected important local features for each modal to construct a cross-modal graph.

\subsection{Deep Cross-Modal Learning based GANs}

Recently, with the development of GANs, many popular deep cross-modal learning methods have been proposed based on GANs. Among these, cross-modal retrieval is a hot research topic attracting tremendous attention. To deal with the problem of cross-modal retrieval, Wang et al.~\cite{DBLP:conf/mm/WangYXHS17} presented ACMR, a novel cross-modal subspace learning method that generates modality-invariant representations by inter-modal adversarial loss. Peng et al.~\cite{DBLP:journals/tomccap/PengQ19} proposed Cross-modal Generative Adversarial Networks (CM-GANs), which consisted of two GAN models. This approach explored inter-modality and intra-modality correlation simultaneously. Li et al. \cite{li2019coupled} proposed an Unsupervised coupled Cycle generative adversarial Hashing networks (UCH), where outer-cycle network is used to learn powerful common representation, and inner-cycle network is explained to generate reliable hash codes. The proposed UCH has achieved excellent performance for the task of cross-modal retrieval. Inspired by zero-shot learning, Xu et al. \cite{xu2019ternary} proposed a novel model called ternary adversarial networks with self-supervision (TANSS). TANSS can employ self-supervised learning to zero-shot cross-modal retrieval problem, which learned the common semantic space with some additional supervision, rather than directly relying on the labels. He et al. \cite{He2017Unsupervised} constructed an unsupervised cross-modal method, named UCAL, which can deal with the cross modal retrieval based on adversarial learning. Specially, UCAL proposed an additional regularization using adversarial learning, which can maximize the correlations between different modalities. Based on transfer learning, Huang et al.~\cite{DBLP:journals/tcyb/HuangPY20} proposed to conduct knowledge transfer from single-modal source domain to cross-modal target domain. They construct a novel model, namely modal-adversarial hybrid transfer network (MHTN), to improve the cross-modal performance. Similar as the studies above, they utilized modal-adversarial consistency loss to learn modality-consistent representations. Wen et al.~\cite{DBLP:conf/icmcs/WenHYL19} developed a cross-modal similarity transferring (CMST) framework to explore the semantic correlations between unpaired items in an unsupervised way. They studied to learn the quantitative single-modal similarities, and transferred them to the common representation subspace. Wang et al.~\cite{DBLP:journals/kbs/WangPHS19} combined autoencoder and adversarial learning to construct a novel multi-view representation learning method, named AdvCAE, which comprehensively utilized the information from all views and enforced the aggregated posteriors of different views to be identical.

Some other deep cross-modal methods have also been proposed to deal with various tasks. Zhang et al. \cite{zhang2018unsupervised} constructed a cross-modal hashing model based on GANs in an unsupervised fashion. They also developed a correlation graph based approach that can capture the underlying manifold structure across multi-modalities. Gupta et al. \cite{gupta2016cross} proposed a novel cross-modal method which can transfer supervision from labeled RGB images to unlabeled depth and optical flow images. Therefore, it can extend various deep CNN architectures to novel imaging modalities without training deep CNNs using a large set of collected images. Dan et al. \cite{Dan2019Semi} proposed a cross-modal image generation model which can deal with semi-supervised problems. It contained one generator and two discriminators and leveraged large amounts of unpaired data. Li et al. \cite{li2020infrared} focused on the problem of person re-identification and introduced an auxiliary X modality which is generated by a lightweight deep network. Then, the network is learned in a self-supervised manner with the labels inherited from visible images. With the help of the auxiliary X modality, cross-modal learning is guided by a carefully designed modality gap constraint, which achieved excellent performance for person re-identification \cite{YangCV3,YangCV4,YangCV5,YangCV6,YangCV7}. Deep Modality Invariant Adversarial Network (DeMIAN) \cite{harada2017deep} aimed to learn modality-invariant representations from paired modalities, while enabling to train a classifier on source modality samples, which can perform well on target modality. To reveal the correlations in human perception of auditory and visual, Chen et al. \cite{chenlele2017deep} attempted to deal with the cross-modal generation task by leveraging GANs. AugGAN \cite{huang2018auggan} served as a structure-aware image-to-image translation network that is capable of handling cross-modal data augmentation. AugGAN can extract structure-aware information through the supervision of a segmentation subtask, which can achieve good performance for object detection.

As discussed above, \cite{DBLP:journals/tomccap/PengQ19} proposed CM-GANs framework that utilized GANs to model cross-modal joint distribution of different modalities. CM-GANs is a cross-modal GAN architecture which is trained by jointly solving the learning problem of two parallel GANs as follows:

\begin{equation}
\min \limits_{G_I,G_T} \max \limits_{D_I,D_T,D_{Ci},D_{Ct}} \mathcal{L}_{GAN_1} \left(G_I,G_T,D_I,D_T\right) +  \mathcal{L}_{GAN_2} \left(G_I,G_T,D_{Ci},D_{Ct}\right)
\label{eq2}
\end{equation}

The intra-modality discriminative model are composed of two sub-nets for image and text, denoted as $D_I$ and $D_T$. Inter-modality discriminative model denoted as $D_C$. $D_{Ci}$ and $D_{Ct}$ represent image pathway and text pathway,  respectively. $G_I$ and $G_T$ represent the generative models for image and text.  A cross-modal adversarial training mechanism is proposed in CM-GANs as Eq.\ref{eq2}, which deployed two kinds of discriminative models to simultaneously conduct intra-modality and inter-modality discrimination. They can mutually boost each other to make the generated common representations more discriminative via the adversarial training process.

\section{Possible Future Direction}
All the above work concentrated on fusing the multi-view information to obtain a better performance than single view counterparts. We state that the multi-modal research may enhance larger weight on \emph{how to solve the problem for each modal by referring to the information from other modalities} for future work.

\begin{figure}
\centering
\includegraphics[width=.8\textwidth]{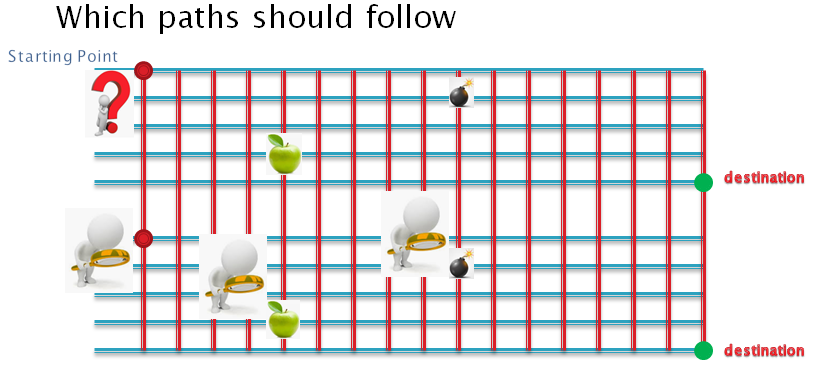}
\caption{Two candidates from two modalities attempt to reach the destinations throughout same network, where they may provide the information to each other to avoid being trapped into bomb, and arrived at the destination. The spatial-temporal multi-modal collaboration are required.}
\label{fig:direction}
\end{figure}

Suppose that we solve the specific problem of path selection over the road network with two modalities. For either of them, the challenges of large complexity or uncertainty may be available. To address the bottleneck, the informative knowledge may be required from other modal, enabling the problem to be feasible. We provide an example in Fig. \ref{fig:direction}, where two candidates (two modalities) aim at reaching the final destinations from the starting points in the context of the same road networks, the challenge lies in which paths should follow. Besides, some paths may be trapped, others may be awarded. One straightforward strategy goes to brute force manner, which, however, is prohibited in term of complexity for large scale networks. Hence, each candidate is expected for the informative knowledge from the other candidate, such as the correct selection for some local network to simultaneously reduce the complexity and uncertainty, to reach the final destination.

\subsubsection{Spatial-Temporal Multi-modal Collaboration}
Based on the example above, one natural question is raised up. That is, how could two candidates help each other? To answer this questions, we need to resolve two basic questions, namely \emph{where} and \emph{when} to collaborate. Specifically, one candidate needs to figure out the right time slot, to guide the other one to capture the information. Here are two major scenarios:

\begin{itemize}
\item If too late, the candidate cannot step back;

\item If too early, the candidate cannot even be aware of the situation unless skip the current context.
\end{itemize}

To effectively solve that, the ideal deep multi-modal models are expected to obtain powerful feature representations and spaces, where the robust model can be trained to achieve the goal. Such motivation may call for interesting research to solve the problem under more practical context.

\section{Conclusions}
We provide a comprehensive overview of the state-of-the-arts regarding Deep Multi-modal data analytics. We claim that the critical issue over existing state-of-the-arts is how to perform the multi-modal collaborations, including adversarial deep multi-modal collaboration, so as to fuse these complementary multi-modal information to be the final output. We also summarize the experimental results of the state-of-the-art deep multi-modal/cross-modal architectures over benchmark multi-modal data sets. A comprehensive review for the practical applications solved by the existing deep multi-modal methods is also offered.

Finally, we propose some future directions, to reveal that multi-modal data analytics may served as the useful information to reconcile different modalities to solve the problem, in terms of expensive complexity, uncertainty etc. The intuitive example is provided for illustration. Hopefully, some future research can be proposed to tackle the problems, and raise up more chances in the field of multi-modal data analytics.
\section{Acknowledgments}
This research is supported by National Natural Science Foundation of China, under grant number of NSFC 61806035 and NSFC U1936217.

All content represents the opinion of the authors, which is not necessarily
shared or endorsed by their respective employers and/or sponsors. 
\bibliographystyle{ACM-Reference-Format}
\bibliography{mybibfile}

\end{document}